# New Algorithms for Computing Field of Vision over 2D Grids

Evan R.M. Debenham and Roberto Solis-Oba

Department of Computer Science, The University of Western Ontario, Canada

## Abstract

*The aim of this paper is to propose new algorithms for Field of Vision (FOV) computation which improve on existing work at high resolutions. FOV refers to the set of locations that are visible from a specific position in a scene of a computer game.*

*We summarize existing algorithms for FOV computation, describe their limitations, and present new algorithms which aim to address these limitations. We first present an algorithm which makes use of spatial data structures in a way which is new for FOV calculation. We then present a novel technique which updates a previously calculated FOV, rather than re-calculating an FOV from scratch.*

*We compare our algorithms to existing FOV algorithms and show they provide substantial improvements to running time. Our algorithms provide the largest improvement over existing FOV algorithms at large grid sizes, thus allowing the possibility of the design of high resolution FOV-based video games.*

## 1. Introduction

### 1.1. Background

A Field of Vision (FOV) is the set of locations that are visible from a specific position in a scene of a computer game. FOV is calculated over a two-dimensional finite grid which is referred to as the FOV grid. An FOV grid partitions a game's environment into rectangular cells. If a cell within this grid contains an object which is vision-blocking (such as a wall, a tree, etc.) then that entire cell is vision-blocking. Games use an FOV grid as it allows them to consider the visibility of a whole region at once, which is significantly less computationally expensive than considering the visibility for every point in the environment. Because of this, the FOV grid may be constructed at a different resolution than the resolution at which the game will be displayed. One grid cell is specified as the source of vision and is referred to as the FOV source cell. An FOV algorithm must determine which cells are visible from the source and which cells are not visible based on the cells that are vision-blocking; the resulting grid with cells set to visible or non-visible is called the FOV.

Figure 1 shows an FOV for a scene of Crypt of the Necrodancer by Brace Yourself Games. The game environment is shown on the left with the FOV grid superimposed in purple. The yellow brick walls block vision from the character located near the bottom left corner. In this example each FOV grid cell is a 48*48 pixel region of the game's display. A data representation of this scene is shown on the right. The FOV source cell is marked with an S, vision-blocking cells are yellow, visible cells are in light brown, and non-visible cells are darkened.

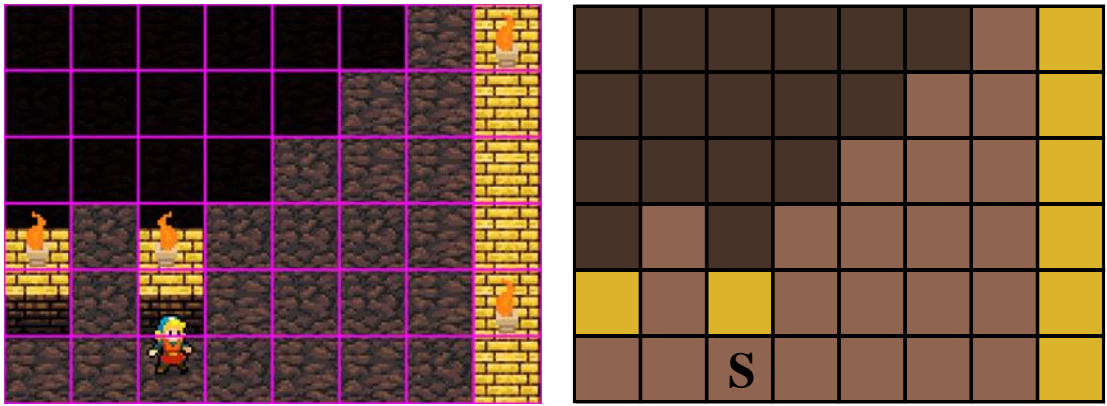

Figure 1. FOV in a game with simple 2D graphics. Left image taken from Crypt of the Necrodancer, by Brace Yourself Games. Right image shows a data representation of the FOV.

Calculating an FOV is useful for computer games with a top-down perspective. In these games the player views the game world from above and thus sees much more of the game world than an individual character inside the game. Top-down games may make use of FOV to provide accurate visibility information to computer-controlled game actors, so that they may then make decisions based on what they can see. Top-down games may also use FOV to convey to a player which areas of the world their character cannot see by visually darkening them. This visual effect is referred to as a fog of war.

FOV is used by a number of popular computer games, such as League of Legends [1, 2] and Defense of the Ancients 2 [3]. The time needed to compute an FOV must be considered when designing games with complex environments, and both of the above games calculate FOV at reduced resolutions in order to improve performance.

Games are expected to render their scenes many times per second in order to give the impression of smooth motion. These rendered scenes are then displayed as frames on a computer screen. Most computer displays support a maximum of 60 frames per second (one new frame roughly every 17ms), but in newer displays this number has increased to up to 240 frames per second (one new frame roughly every 4ms). The faster and more consistently a game is able to render its scenes the smoother the display of the game will be for the user. A game might not need to calculate the FOV every time a scene is rendered, but it must be able to calculate it fast enough such that rendering is not delayed. Games also have many different processes that must share system resources. Because of these conditions, an FOV algorithm must complete extremely quickly so as not to delay rendering or starve other processes of system resources. We show that existing FOV algorithms do not scale well with grid size and have inadequate performance as a result.

## 1.2. Existing FOV Algorithms and Related Work

All existing FOV algorithms make use of rays cast from the FOV source. One such algorithm is Mass Ray FOV, which casts a ray from the center of the FOV source cell to the center of each cell in the FOV grid, and if that ray does not intersect any vision-blocking cells then that ray's destination cell is marked as visible. Mass ray FOV has very poor performance because it casts as many rays are there are cells in the grid. Another algorithm based on direct ray casting is Perimeter Ray FOV. This algorithm casts a ray to the center of every cell on the perimeter of the grid and for each ray it sets to visible all cells that the ray touches before the ray finds a vision-blocking cell. While Perimeter Ray FOV does cast fewer rays than Mass Ray FOV, all algorithms which directly cast rays to specific grid cells may cast more rays than necessary because they cast a fixed number of rays regardless of the terrain.

More intelligent FOV algorithms selectively cast visibility rays to the corners of vision blocking cells and use these rays to determine the boundaries of the visible space. Recursive Shadowcasting by Björn Bergström [4] is the most popular publicly available algorithm which is based on this selective ray casting approach. Recursive Shadowcasting starts by initializing all grid cells to not visible. It then splits the FOV grid into eight octants centered on the FOV source and traverses the cells within each octant as shown in Figure 2. This traversal occurs within each octant by rows or columns in ascending order of distance from the FOV source. As a cell is traversed, its visibility status is set to visible. However, when a vision-blocking cell is encountered, an octant is split into two smaller sub-octants which are bounded by rays cast from the FOV source to the corners of the vision-blocking cell. The cell traversals are then continued within each sub-octant.

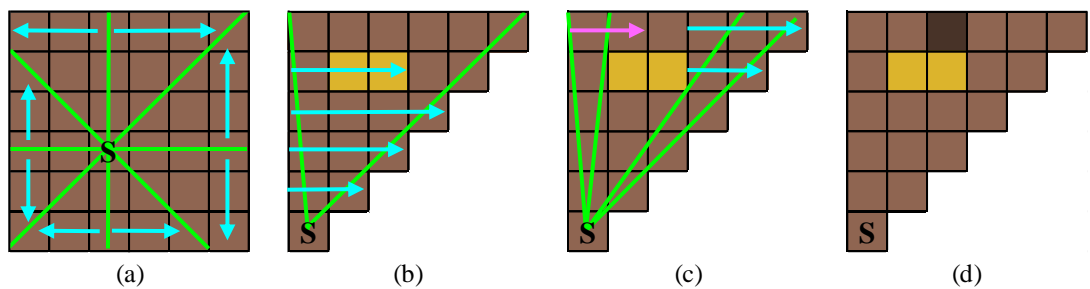

Figure 2. (a) An FOV grid split into octants with row/column traversal shown. (b) Traversal of an octant until vision-blocking cells are encountered. (c) Traversal of sub-octants after the traversal is split by the vision-blocking cells. (d) The resulting FOV.

Permissive FOV by Jonathon Duerig [5] is another popular FOV algorithm which selectively casts rays. Recursive Shadowcasting and Permissive FOV are efficient at low grid sizes but become slow as grid size increases because they perform a relatively large number of operations per-cell. It is important to note that FOV grids can be made of tens of thousands of cells. For an algorithm to improve on the performance of selective ray casting, it must determine cell visibility and set cell visibility statuses in a more efficient manner.

There are some problems in computer games that involve determining visibility information, like determining how light and shadow interact with the objects in a scene, and how to ensure that parts of objects which are not visible are not rendered. Techniques that have been designed to address these problems, such as shadow mapping [6], shadow volumes [7], real-time ray tracing [8], z-buffering [9], the painter's algorithm [10], frustrum culling [11], and portal-based occlusion culling [12] cannot be used to calculate an FOV.

The rest of the paper is organized in the following manner. In Section 2 we propose a new FOV algorithm named Rectangle-Based FOV, which represents vision-blocking cells in a very compact and efficient way by using rectangles. In Section 3 we propose a second new algorithm named FOV Update, which adjusts a previously calculated FOV instead of calculating an FOV from scratch. In Section 4 we compare both of these algorithms to Recursive Shadowcasting (as it is known to have the best performance among existing FOV algorithms [13, 14]) and determined that our algorithms offer superior performance to Recursive Shadowcasting when the grid size becomes large. In Section 5 we summarize our results and make recommendations to implementors who may wish to use our algorithms.

## 2. RECTANGLE-BASED FIELD OF VISION

### 2.1. Representing Vision-Blocking Cells with Rectangles

In computer games, the FOV usually needs to be calculated every time the FOV source moves, but the game environment changes rather infrequently. Because of this, it is possible to pre-process the environment and represent the vision blocking cells in a compact and efficient manner using rectangles. This efficient representation can then be used for many FOV calculations, and only needs to be updated if the game environment changes. The performance of an FOV algorithm which uses such a compact representation of vision-blocking cells is less dependent on the size of the FOV grid than selective ray casting algorithms.

We use rectangles to represent groups of vision-blocking cells, this allows us to compactly represent the vision-blocking terrain in a game environment. The rectangle-based representation of vision-blocking cells is created with the following process. Adjacent vision blocking cells are first grouped together into rectilinear regions. These rectilinear regions are then split into a minimal number of rectangles using the algorithm in [15]. The rectangles which represent vision-blocking cells on a grid are stored in a quadtree [16] which allows us to rapidly access them, instead of having to search for them within the grid. Quadtrees are simple to build and update, which is important if the game environment does change.

A quadtree stores rectangles by hierarchically dividing the FOV grid into successively smaller quadrants. The root node of the quadtree represents the entire FOV grid. Each internal node (including the root) has exactly four children, each representing one quadrant of the space represented by their parent. When the region that a node represents contains fewer than a predefined number N of rectangles, that node is made into a leaf and it stores all rectangles that intersect the region that it represents (see Figure 3). Note that a rectangle may be stored in multiple leaf nodes.

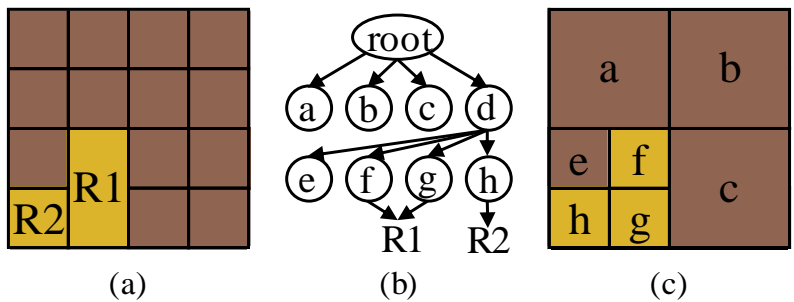

Figure 3. (a) A grid containing two rectangles R1 and R2. (b) The quadtree that represents this grid, with N = 1. (c) The region represented by each leaf node of the quadtree

We consider a cell to be visible if any point within that cell can be seen from the FOV source. We use this definition of visibility as it is the most commonly used in computer games, and it matches the definition of visibility used by Recursive Shadowcasting. However, our algorithm can be adapted to use other definitions of visibility as well, such as a cell being visible only if its center can be seen by the FOV source.

When determining what areas of the FOV grid are occluded behind a rectangle, we use two of the rectangle's four vertices, and refer to these as a rectangle's relevant points for determining visibility. The relevant points are always the two vertices of the rectangle which are farthest apart from each other among the vertices which are visible to the FOV source when considering that rectangle in isolation (see Figure 4).

We cast a ray from each relevant point in the opposite direction of the FOV source. The area between the rectangle and these two rays contains the area occluded behind that rectangle. Note that this area does not include the rays themselves, as they are considered to be visible from the FOV source. A cell is occluded by a rectangle if it is entirely within the area (see Figure 4).

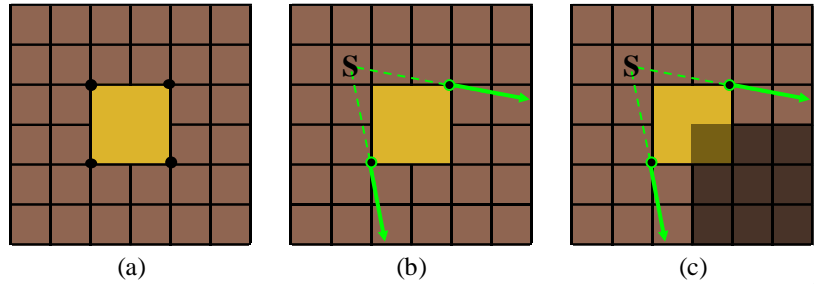

Figure 4. (a) A rectangle with its four vertices highlighted. (b) The two relevant points and casted rays for this rectangle. (c) The cells occluded by the rectangle are darkened.

## 2.2. Calculating an FOV using Rectangles

We process each rectangle as specified below in order to assign visibility statuses to the cells of the FOV grid. However, first we note that we cannot process each rectangle in isolation. If two rectangles R1 and R2 have (part of) a common side, there might be some cells which are not visible from the FOV source but which are not fully contained in the occluded region of either rectangle. If R1 and R2 are processed independently then these cells might be incorrectly labelled as visible. Figure 5 gives an example of this: Cells i and ii are not fully occluded by either rectangle R1 or R2 but they are not visible to the FOV source. Note that i and ii are not considered to be occluded by R1 because one of the rays cast from R1 touches the corners of i and ii. Figure 5(a) and (b) show the individually occluded regions for R1 and R2 respectively, and Figure 5(c) shows how these combined occluded areas do not result in a correct FOV.

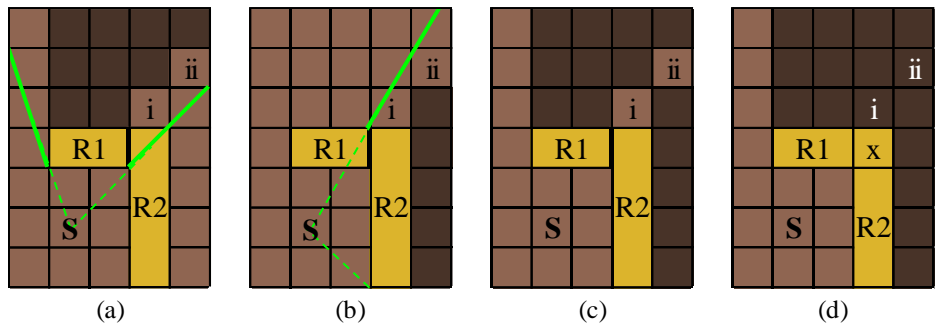

Figure 5. An environment where rectangles R1 and R2 are adjacent. (a) The region occluded by R1. (b) The region occluded by R2. (c) cells i and ii are incorrectly labelled as visible. (d) Correct labelling of the cells.

We address this issue by slightly increasing the size of one of the two adjacent rectangles, as this will ensure that all cells are correctly set to not visible. For each relevant point P of R1, we check if P also belongs to another rectangle. This check can be efficiently performed using the quadtree: We recursively traverse the nodes of the quadtree which represent regions containing P until we reach a leaf node. Upon reaching a leaf node we check all rectangles stored in it (except R1) to see if any of them contains P. If some rectangle R2 contains P, we check if R2 occludes P. If P is not occluded, we extend the size of R1 by one row or column so that it overlaps with R2. In the example in Figure 5(d), R1 is extended by one column to the right into R2, so the rectangles now overlap at cell x. This results in cells i and ii being correctly set to not visible as the extended R1 occludes them.

After rectangles have been extended (if needed), the visibility statuses of the grid cells are assigned: Initially each cell is assigned a status of visible; then we process the rectangles one by one. For each row of the FOV grid that is within the occluded region of a rectangle, the algorithm calculates the leftmost and rightmost cells which are fully within the occluded region; the algorithm then traverses the row from this leftmost cell to the rightmost cell, setting every cell traversed to not visible. Once this process is repeated for every rectangle, every grid cell that is occluded by a rectangle will have its visibility status set to not visible, while all other cells will remain as visible.

## 2.3. Optimizations to Rectangle-Based FOV

The algorithm described in Section 2.2 will correctly calculate an FOV but it may perform redundant work. In this section we describe two performance optimizations that reduce the amount of work performed by the algorithm. Firstly, when the rectangle being processed is partially or totally occluded behind other rectangles, the regions that these rectangles occlude overlap. As we set the cell visibility statuses within these occluded regions, some cells might be set to not visible multiple times. We address this by reducing the size of the rectangle we are processing such that it is no longer occluded by other rectangles.

If we process rectangles in an order such that a rectangle is processed after any rectangles which occlude it, then when processing a rectangle we could use any partially computed visibility information to determine whether a rectangle is (partially) occluded. We can traverse the quadtree to efficiently arrange the rectangles so that many of them are processed after any rectangles that (partially) occlude them, as we explain below.

We define the distance between the FOV source and a rectangular region of the FOV grid as the distance between the FOV source and the closest point within that region. Starting from the root, we recursively process the children of each internal node of the quadtree in order from closest to furthest distance from the regions represented by the child nodes to the FOV source. For a leaf node, we process its rectangles in order from closest to furthest from the FOV source. Since a rectangle can be stored in more than one node, we only process a rectangle the first time it is encountered. Note that our ordering does not guarantee that a rectangle will always be processed after rectangles which occlude it, but for many rectangles this will be the case.

Because we determine the entire area occluded by a rectangle at once, we can take advantage of spatial locality in order to set visibility statuses efficiently. Spatial locality refers to a property of computer memory where accessing data sequentially is substantially faster than accessing it in a random order. Memory has this property due to the nature of the CPU cache, which stores recently accessed data and the data which is adjacent to it. By accessing data in a sequential manner, we take advantage of the cache storing adjacent data, and so are able to access data from the cache instead of main memory and we can set cell visibility statuses quickly.

## 2.4. The Rectangle-Based FOV Algorithm

First the vision-blocking terrain must be converted to a quadtree of rectangles. The steps for this process are summarized in pseudocode below:

**Algorithm:** Vision-Blocking Cells to Rectangles(G, N)
Input: FOV grid G, integer N specifying the maximum number or rectangles in a leaf node.
Result: The root of a quadtree which contains the vision-blocking rectangles of G.

Let L be an empty list of rectangles.
**for each** rectilinear region E of vision-blocking cells **in** G:
    Dissect E into rectangles (as described in Section 2.1) and add those rectangles to L.
Let Q be a quadtree node representing the entire grid G. //Q is the root of the quadtree
**for each** rectangle R **in** L:
    Add R to every leaf node of the quadtree representing a region that intersects R.
    **while** there is a leaf node P which has more than N rectangles:
        Convert P into an internal node with 4 children, each representing one quadrant of P.
        Move each rectangle r in P into every one of P's children which intersect r.
**return** Q

Once the quadtree has been constructed and all cells of the grid have been initialized as visible, the rectangle-based FOV algorithm is used to compute the FOV from a given source cell. The steps of the Rectangle-Based FOV algorithm are summarized in pseudocode below:

**Algorithm:** Rectangle-Based FOV(N, S, G)
Input: Quadtree node n, FOV source cell S, FOV grid G.
When first called: n will be the root node of the quadtree and all cells in G are set to visible.
Result: Cells in G which are not visible from S are set to not visible.

**if** n is a leaf node **then:**
    **for each** rectangle R **in** n, from closest to farthest from S:
        **if** R has not already been processed **then**
            Extend R if needed, as described in Section 2.2.
            Shrink R if needed, as described in Section 2.3.
            Let E be the region occluded behind R.
            **for each** row X **of** the grid G that intersects E:
                Set to not visible the cells in X contained in E, from left to right.
            Mark R as processed

    **else**:
        **for each** child node C **of** n, from closest to farthest from S:
            Rectangle-Based FOV(C, S, G)

## 3. UPDATING AN EXISTING FIELD OF VISION

All FOV algorithms we have discussed so far calculate the FOV from scratch and require the cells of the FOV grid to be initialized as either all non-visible or all visible. This initialization significantly affects performance at large grid sizes. Additionally, these FOV algorithms completely discard the previously calculated FOV when they reset the visibility information stored in the grid cells.

Our second new FOV algorithm uses the previously calculated FOV rather than discarding it. This improves performance as this algorithm does not need to clear the grid and so it will need to assign fewer cell visibility statuses. Updating an FOV is possible because an FOV often needs to be re-calculated when the FOV source moves to an adjacent grid cell. Hence, it is likely that most FOV grid cells will have the same visibility status in both FOV calculations. Therefore, we may be able to compute the new FOV more efficiently if we update a previously calculated FOV instead of calculating it from scratch.

### 3.1. Cones of Changing Visibility

Updating a previously calculated FOV is conceptually more complex than calculating an FOV from scratch. When updating an FOV two FOV sources must be considered: S1, the source for which the FOV was previously calculated, and S2, the new source for which the FOV is being updated. Each vision-blocking rectangle has two relevant points when considering S1, and two relevant points when considering S2. A ray is cast from each relevant point in the opposite direction of each FOV source. A ray is considered to be visible from the FOV source from which it is cast. The four rays cast from a rectangle are grouped to form two ray pairs such that:

- Each pair is made of one ray directed away from S1 and one ray directed away from S2.

- Two of the rays in a pair share a relevant point and the other two rays either share a relevant point or lie on a common side of the rectangle.

The area between two rays in a pair, and possibly the common side of the rectangle containing their relevant points, is called a cone (see Figure 6(c)). When considered in isolation, a cone contains space which is occluded behind its rectangle from either S1 or S2, but not both. The two cones cast from a rectangle represent space where the visibility status may change when updating the FOV. The point within a cone that is closest to both S1 and S2 is called the origin of the cone. Figure 6 gives an example of how cones are created. From the preceding discussion we obtain the following observations.

Proposition 1: The set of cones cast from all rectangles represent the only space where the visibility status may change when the FOV sources moves from S1 to S2.

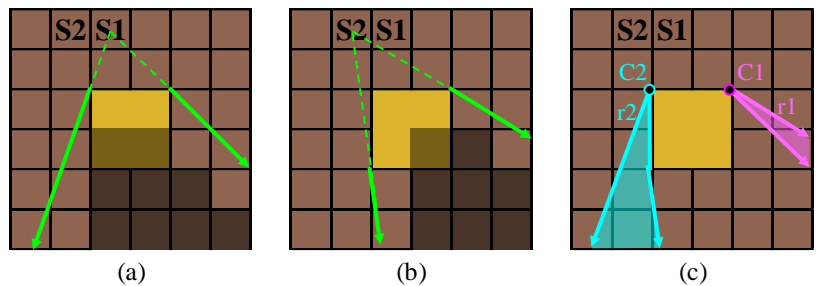

Figure 6. (a) Rays and cell visibility for S1. (b) Rays and cell visibility for S2. (c) The rays pairs forming two cones C1 and C2, with their origin points marked with a circle. The rays forming C1 are cast from the same point, while the rays forming C2 are cast from different points.

When considering a cone in isolation and the rectangle from which that cone is cast, one ray will always be adjacent to space which is visible to both FOV sources. This ray is called an outer ray, while the other is referred to as a cone's inner ray. For example, in Figure 6(c), the two rays r1 and r2 which are further away from the rectangle are the outer rays, and the two other rays are the inner rays. The outer ray of a cone is the only part of it which is visible to both FOV sources, this property will be used in Section 3.2.

We classify cones into three categories based on the visibility of their origin. If the origin of a cone is visible to both S1 and S2, that cone is fully visible. If the origin of a cone is visible from S1 or S2, but not both, that cone is said to be transitioning visible. If the origin of a cone is neither visible from S1 nor from S2, that cone is said to be not visible. See Figure 7.

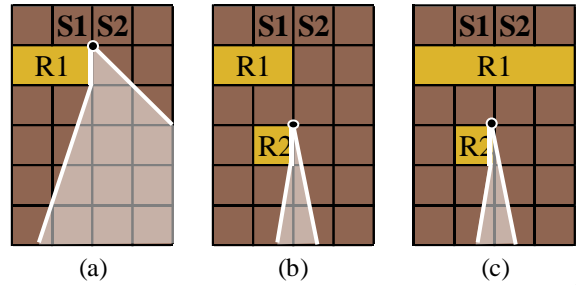

Figure 7. (a) A fully visible cone. (b) A transitioning visible cone. (c) A not visible cone.

## 3.2. Inverting Cones to Update Cell Visibility

Note that from here onward we assume that the FOV source always moves from its current location to an adjacent cell of the grid, i.e. S1 and S2 are assumed to be at the centers of grid cells which share a side. This restriction means that the FOV source cannot move through or around a vision blocking rectangle in a single FOV update. The less common computer game scenario of the FOV source moving to a non-adjacent grid cell can be addressed by performing several FOV updates in which the FOV moves to adjacent cells, or by re-calculating the FOV from scratch if the FOV source moves a large distance.

Inverting a cone C means to invert the visibility status of all grid cells which intersect C, but do not intersect C's outer ray. Recall that we define a cell as visible to an FOV source if any part of it is visible, even the very edges. This definition is consistent with the definition of visibility used by Recursive Shadowcasting and Rectangle FOV. Figure 8 shows an example of cells whose visibility status change when a cone is inverted.

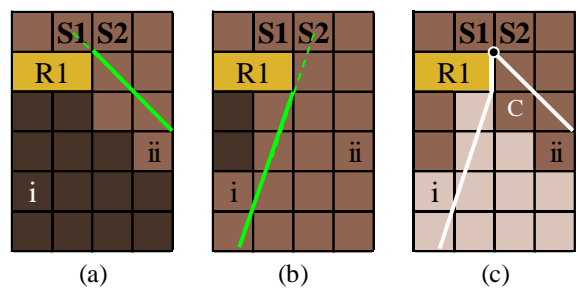

Figure 8. (a) Cells not visible from S1 are darkened. (b) Cells not visible from S2 are darkened. (c) The visibility status of the cells highlighted in white is changed when cone C is inverted. Note that cell i just touches the inner ray of C, so its visibility status changes as it is visible only from S2. Cell ii just touches the outer ray of C, and so it can be seen by both sources.

We present a series of lemmas below, which will allow us to show that to update the FOV we only need to invert all fully visible cones. Note that the proofs for the lemmas, theorems, and corollaries have been omitted due to space limitations.

Lemma 1: If the origin of a cone C2 is within another cone C1, then C2 is entirely within C1.

Lemma 2: Any part of a non-visible cone C that is visible to S1 or S2 must be contained within another cone.

Theorem 1: The FOV can be correctly updated when the FOV source moves from S1 to S2 by only addressing cells which intersect the fully visible cones.

## 3.3. Handling Intersections of Rectangles and Fully Visible Cones

Consider a fully visible cone C and the rectangle R from which it is cast. When considering C and R in isolation, the outer ray of C is visible to S1 and S2, and the rest of C is only visible to one of the FOV sources. However, in an environment with many rectangles, some rectangles may intersect C, causing some parts of C to be visible to neither FOV source. If any rectangle Ri intersects C, then C is shrunk as described below, such that we remove from it any regions occluded from both FOV sources by Ri.

Before it is shrunk, a fully visible cone C is a region C′ which is bounded by an outer ray, an inner ray, and possibly a side of R. We call this region C′ the expanded cone of C. Note that if C is not shrunk then C = C′, otherwise C is contained in C′.

Shrinking C is accomplished by shortening the existing rays of C′ and/or by adding new rays. There are three cases of intersection to consider between C and a single rectangle Ri:

- If Ri intersects both rays of C′, then both rays of C′ are shortened so that they end as they intersect Ri. C is bounded by these new shorter rays, Ri, and possibly a side of R.

- If Ri intersects only one ray r1 of C′, then r1 is shortened so that it ends where it intersects Ri. A new ray r2 is then added to C, which is directed away from the same FOV source as the inner ray of C′; r2 is cast from the relevant point of Ri which is inside of C′. C is then bounded by the shortened ray r1, the other ray of C′, Ri, r2, and possibly a side of R.

- If Ri does not intersect either ray of C′, then Ri is completely within C′ and it effectively splits C′ in a manner which is similar to Recursive Shadowcasting. This requires adding two rays to C; one ray is cast from each relevant point of Ri with respect to the FOV source from which the inner ray of C′ is cast, and both rays are directed away from that FOV source. In this case C is bounded by the rays of C′, Ri, the two rays cast from Ri, and possibly a side of R.

Note that for the purposes of cone shrinking, a ray of C is considered to intersect Ri even if it is colinear to a side of Ri, or it if contains one of Ri's vertices.

If multiple rectangles intersect C′, the cone is shrunk using the same process described above, considering the intersecting rectangles in increasing order of distance from the FOV sources. Figure 9 gives an example of a cone intersecting multiple rectangles. Rectangle R5 intersects two rays as in the first case, rectangles R3 and R4 intersect one ray as in the second case, and rectangle R2 intersects neither rays as in the third case.

An important note to keep in mind is that if any rectangle intersects the outer ray of C′, then the outer ray is shortened as explained above. Even after being shrunk, the outer ray remains the only part of C which is potentially visible to both FOV sources.

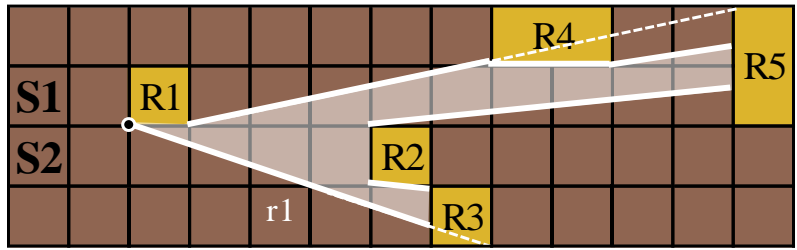

Figure 9. An example of a cone C intersecting multiple rectangles. The solid line and colored regions represent C, while the dotted lines represent the rays of the expanded cone C′. Note that the outer ray r1 is shortened due to intersection with R3. All rays except for r1 are cast from S2.

When a cone C, the rectangle R from which C is cast, and all rectangles intersecting C are considered in isolation, each cell intersecting C but not intersecting its outer ray is visible to only one FOV source, while a cell intersecting the outer ray of C is visible to S1 and S2. However, additional rectangles not intersecting C may occlude parts of C from S1 or S2; these occluded parts then must be within the intersection of C and other cones. Hence, from the above discussion we derive the following observation.

Proposition 2: Each cell g which intersects a fully visible cone C but not its outer ray is either:

- Visible to only one FOV source.

- Not visible to either source. In this case g intersects the intersection of C with another fully visible cone C1, but g does not intersect the outer ray of C1.

Each cell g intersecting the outer ray of a fully visible cone C is either:

- Visible to both FOV sources.

- Visible to only one source. In this case g intersects the intersection of C with another fully visibly cone C1, but g does not intersect the outer ray of C1.

Lemma 3: If the outer ray of a cone C is cast from an FOV source S, no part of C, except possibly its outer ray, is visible from S.

We define the inner intersection of two intersecting cones as the intersection of the two cones excluding their outer rays.

Lemma 4: The grid cells completely contained within the inner intersection of two fully visible cones C1 and C2 do not change visibility status when the FOV source moves from S1 to S2.

Corollary 1: Three or more fully visible cones cannot have a non-empty inner intersection.

Theorem 2: Inverting all fully visible cones in any order will correctly update the FOV when the FOV source moves from S1 to S2.

In order to use Theorem 1 to efficiently update the FOV when the FOV source moves, we must be able to quickly determine the visibility of the origin of each cone. This can be accomplished using a line of sight check: If a straight line traced from a cone's origin to an FOV source intersects any rectangles, then that cone's origin is not visible from that FOV source.

We can efficiently identify the visibility status of the origins of cones by ordering the cones such that a fully visible cone always precedes all transitioning visible cones which are inside of it. The quadtree which stores the vision-blocking rectangles can be used to efficiently create the above ordering using a process similar to how rectangles were ordered in Section 2.2. By inverting the cones in this order, we ensure that any fully visible cone will be inverted before considering any transitioning visible cones within it. This enables us to identify and then ignore transitioning visible cones without using lines of sight.

### 3.4. Inverting Fully Visible Cones

After ordering the cones as described above, we invert the first fully visible cone in the order.

Consider a cone C and the line b that bisects it. If the slope of b is between $-\pi/4$ and $\pi/4$, or between $3\pi/4$ and $5\pi/4$, then C is said to be primarily horizontal. Otherwise C is said to be primarily vertical. To invert a fully visible cone C, we invert the visibility status of the cells which intersect it, except for those cells which intersect its outer ray. If C is primarily horizontal we invert the visibility of each column of grid cells intersecting the cone, one at a time from closest to furthest to the cone's origin. If C is primarily vertical, then we process the cells by rows. We process cells in this order to ensure that if a given cell is part of a vision blocking rectangle R, it will be processed before any of the cells in further rows or columns which may be occluded by R. Below we explain how to process columns of cells of a primarily horizontal cone; a similar process can be used to process rows of cells of a primarily vertical cone.

When processing a column of cells within a cone C, we invert all cells of that column which intersect C, except for any cells which intersect C's outer ray. After the visibility status for all cells in a column have been inverted, we check if any rectangle R intersects C at that column, as then R would occlude cells in further columns. If an intersection is found, the cone C is shrunk as described in Section 3.3.

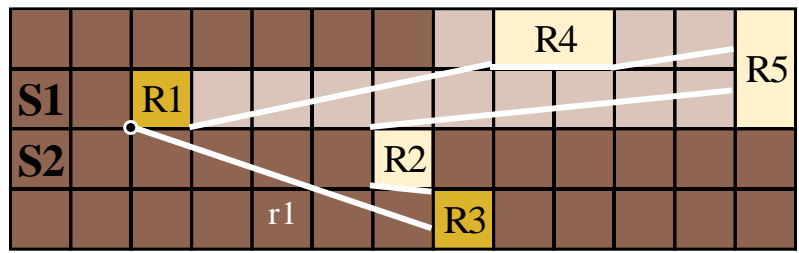

Figure 10: An example of cell visibility inversion for the primarily horizontal cone shown in Figure 9. Cells which have their visibility status inverted (highlighted in white), are all cells which intersect the cone but do not intersect the outer ray r1. Note that the cells within R2, R4, and R5 have their visibility status inverted.

Note that while checking for intersection between a rectangle R and a cone C, we can also quickly identify any cones which are inside of C by storing the vertices of R which are inside of C in a hash table H. If any of those vertices are later found to be the origin of a cone, we know that such a cone must be inside of C and therefore it must be transitioning visible.

After completing the first cone inversion as described above, we iterate over the remaining cones in the aforementioned order. If the origin of a cone is in the hash table H, we know that such a cone is transitioning visible and hence we discard it. If the origin of a cone is not in H, we check if that cone is fully visible or not visible. If the cone is fully visible it is inverted as described above, otherwise it is skipped. After iterating through every cone in this manner, we will have inverted only the cells which are visible from one FOV source, and therefore we will have correctly updated the FOV.

### 3.5. The FOV Update Algorithm

The steps of the FOV Update algorithm are summarized in pseudocode below. We assume that an FOV has already been calculated and a quadtree of rectangles has been created:

**Algorithm:** FOV Update (S1, S2, G, Q)
Input: Grid cells S1 & S2, grid G containing FOV from S1, quadtree Q of rectangles
Result: The grid G will contain the FOV from S2

Let T be the set of all cones cast from the rectangles in Q.
Let H be an empty hash table.
**for each** cone C **in** T, sorted as described at the end of Section 3.3:
    **if** C's origin is not in H **then**:
        **if** the line traced from C's origin to S1 does not intersect any rectangles in Q **then:**
            Invert the cells of G within C as described in Section 3.4.
            Store in H the relevant points of the rectangles in L that are inside of C.

## 4. EXPERIMENTAL EVALUATION

We present an experimental evaluation of Recursive Shadowcasting, Rectangle-Based FOV, and FOV Update. All tests were run a computer with an Intel Xeon E5-2683 processor and 24 gigabytes of system memory. Our algorithm implementations were written in C++, compiled using GCC 8.3.1, and run under Linux Kernel 5.6.8.

We show test results for four environments which are meant to emulate terrain which may appear in a computer game:

**Environment 1:** A fixed indoors environment made of 160 rectangles, with 36 square rooms connected by 74 corridors. This environment is constructed such that there is never an alignment of corridors which would allow the FOV source to see across many rooms. This is an enclosed environment where many cells and rectangles/cones will be occluded.

**Environment 2:** A randomized environment where 200 rectangles of random sizes are placed at random positions on the FOV grid. This simulates a less "structured" environment, such as a forest. Each rectangle has a uniformly distributed random width and height between one and six cells. The position of each rectangle is chosen uniformly at random from all locations that do not intersect another rectangle.

**Environment 3:** A randomized environment where 200 rectangles of random sizes are densely grouped around the center of the FOV grid and fewer rectangles appear further from the center. This simulates a more organized environment, such as a town. Each rectangle has a uniformly distributed random width and height between one and six cells. The position of each rectangle is chosen using a random distribution which results in more positions near the center of the grid.

**Environment 4:** A fixed environment which uses 300 rectangles to emulate the visibility grid used in League of Legends [1]. This tests the FOV algorithms using an environment taken from an existing game that includes a mixture of enclosed spaces and large open pathways.
For the four above environments (shown in Figure 11) we tested the algorithms with 25 randomly generated paths of 100 cells each. Each path was constructed by randomly selecting a non-vision blocking starting cell and a random direction. The starting cell and selected direction define a ray. Cells which intersect this ray were added to the path in ascending order of distance from the starting cell. If the next cell to add to the path belongs to a vision-blocking rectangle, a new random direction was generated and further cells were added to the path using that new direction. This continued until the path contained 100 cells.

The FOV was calculated for each cell along the path in the order in which the cells appear in the path. We used paths in these test environments to mimic scenarios arising in computer games, where the FOV source moves through the grid following a game character. In the case of FOV Update, for each path we first computed an initial FOV using the Rectangle-Based FOV algorithm, and then measured the running time of updating the FOV for every position of the FOV source along the path. Each test environment uses a fixed number of rectangles; as the grid size increases the sizes of the rectangles is increased by the same proportion.

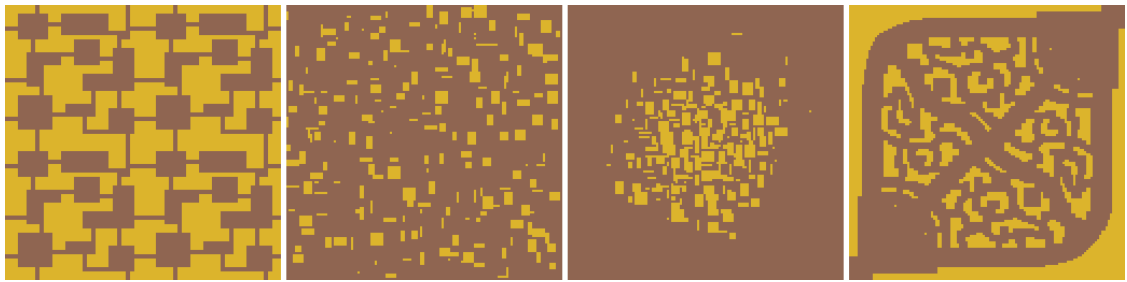

Figure 11. Environments 1, 2, 3, and 4 on a grid of size 128*128.

Table 1. Running times for Environment 1.
Smallest mean running times and smallest standard deviations are highlighted in blue.

|            | Shadow |           | Rectangle |           | Update |           |
|------------|--------|-----------|-----------|-----------|--------|-----------|
| Grid Size  | Mean   | Std. Dev. | Mean      | Std. Dev. | Mean   | Std. Dev. |
| 128*128    | 6.5 μs | 1 μs      | 205 μs    | 20 μs     | 170 μs | 24 μs     |
| 256*256    | 21 μs  | 3 μs      | 259 μs    | 26 μs     | 174 μs | 25 μs     |
| 512*512    | 80 μs  | 14 μs     | 401 μs    | 39 μs     | 188 μs | 27 μs     |
| 1024*1024  | 290 μs | 43 μs     | 774 μs    | 68 μs     | 204 μs | 46 μs     |
| 2048*2048  | 1,342 μs | 278 μs  | 2,001 μs  | 163 μs    | 249 μs | 77 μs     |
| 4096*4096  | 6,665 μs | 1473 μs | 10,269 μs | 765 μs    | 356 μs | 140 μs    |

Environment 1 is designed to have a relatively low number of visible cells from any FOV source position. The low number of visible cells is an advantage for Recursive Shadowcasting and a disadvantage for Rectangle FOV. Rectangle FOV is only about 50% slower that Recursive Shadowcasting at large grid sizes, despite having to assign many more cell visibility statuses, as Rectangle FOV starts with all cells set to visible. This shows that when compared to Recursive Shadowcasting, Rectangle FOV's more efficient method of cell assignment makes a very significant difference to running time, even in environments where Rectangle FOV has to assign many more cell visibility statuses than Recursive Shadowcasting.

The low number of visible cells in this environment is also an advantage to FOV Update, as it results in a high number of non-visible cones and a low number of cells which change visibility status as the FOV source moves. Because of this, FOV Update is faster than Rectangle FOV and it is up to 20 times faster than Recursive Shadowcasting at large grid sizes. FOV update's running time is not strongly affected by grid size in this case due to the low number of cell visibility assignments that it needs to make.

The running times for Environment 1 have the lowest standard deviations of Environments 1-4. This is expected as Environment 1 has a low variance in the number of cells visible from any point, as the shape of the environment ensures that only one room can be fully visible regardless of where the FOV source is positioned. Because of this, the running times of the algorithms are not as strongly affected by the position of the FOV source.

Table 2. Running times for Environment 2.

| Grid Size | Shadow Mean | Shadow Std. Dev. | Rectangle Mean | Rectangle Std. Dev. | Update Mean | Update Std. Dev. |
|---|---|---|---|---|---|---|
| 128*128 | 17 μs | 6.5 μs | 300 μs | 49 μs | 468 μs | 137 μs |
| 256*256 | 54 μs | 16 μs | 358 μs | 52 μs | 504 μs | 135 μs |
| 512*512 | 201 μs | 53 μs | 494 μs | 77 μs | 595 μs | 152 μs |
| 1024*1024 | 777 μs | 289 μs | 943 μs | 172 μs | 763 μs | 243 μs |
| 2048*2048 | 3,898 μs | 1,747 μs | 2,176 μs | 277 μs | 1,073 μs | 366 μs |
| 4096*4096 | 19,345 μs | 8,426 μs | 7,347 μs | 1,059 μs | 1,863 μs | 821 μs |

Recursive Shadowcasting has larger running times in Environment 2 when compared to Environment 1, as Environment 2 has many more visible cells. Due to this increased running time, it is slower than Rectangle FOV when grid size is large. When comparing Rectangle FOV's running times between Environments 1 and 2 the running times for Environment 2 are slower at most grid sizes, but are faster at grid size 4096*4096. At lower grid sizes Rectangle FOV's running time is primarily determined by the number of rectangles, and so its running time is slower than in Environment 1 as more rectangles are present in Environment 2. At 4096*4096 Rectangle FOV's running time is primarily determined by its efficiency of assigning cell visibility statuses, and so it is faster than in Environment 1 as fewer cell visibility statuses need to be assigned in Environment 2.

Many more cells may change visibility status in Environment 2 than in Environment 1 when the FOV source moves, and so FOV Update's running time is more significantly affected by grid size than in the previous environment. FOV Update is much faster than the other two algorithms as grid size becomes large, as it changes the visibility status of fewer cells.

Table 3. Running times for Environment 3.

| Grid Size | Shadow Mean | Shadow Std. Dev. | Rectangle Mean | Rectangle Std. Dev. | Update Mean | Update Std. Dev. |
|---|---|---|---|---|---|---|
| 128*128 | 25 μs | 9.7 μs | 272 μs | 35 μs | 471 μs | 138 μs |
| 256*256 | 83 μs | 35 μs | 314 μs | 43 μs | 466 μs | 142 μs |
| 512*512 | 343 μs | 169 μs | 431 μs | 64 μs | 489 μs | 146 μs |
| 1024*1024 | 2,132 μs | 809 μs | 832 μs | 117 μs | 676 μs | 173 μs |
| 2048*2048 | 11,529 μs | 5,592 μs | 2,072 μs | 226 μs | 969 μs | 269 μs |
| 4096*4096 | 46,203 μs | 25,962 μs | 6,710 μs | 1,007 μs | 1,331 μs | 539 μs |

The running times of Recursive Shadowcasting are much higher for Environment 3 than for Environment 2 because the clustering of the vision blocking rectangles results in a high number of visible cells when the FOV source is not in the center of the grid, which causes Recursive Shadowcasting to calculate the visibility status of many cells. This also explains the high standard deviation of Recursive Shadowcasting, as visibility is low if the FOV is near the center of the grid, and high otherwise. FOV Update's faster running times here than in Environment 2 are due to the clustering increasing the number of rectangles that occlude each other. This reduces the number of cones that the algorithm needs to process.

Table 4. Running times for Environment 4.

| Grid Size | Shadow | | Rectangle | | Update | |
|---|---|---|---|---|---|---|
| | Mean | Std. Dev. | Mean | Std. Dev. | Mean | Std. Dev. |
| 128*128 | 13 μs | 6.5 μs | 403 μs | 57 μs | 558 μs | 220 μs |
| 256*256 | 46 μs | 24 μs | 482 μs | 78 μs | 566 μs | 223 μs |
| 512*512 | 163 μs | 75 μs | 656 μs | 100 μs | 590 μs | 219 μs |
| 1024*1024 | 844 μs | 468 μs | 1,173 μs | 210 μs | 687 μs | 328 μs |
| 2048*2048 | 4,157 μs | 2,780 μs | 2,643 μs | 472 μs | 802 μs | 432 μs |
| 4096*4096 | 22,007 μs | 13,698 μs | 8,692 μs | 1,724 μs | 1,247 μs | 765 μs |

Despite Environment 4 being more enclosed than Environments 2 and 3, Recursive Shadowcasting still performs poorly here. Because the large open pathways in this environment can result in a high number of visible cells. Recursive Shadowcasting's running time is primarily determined by the number of visible cells. This also explains the high standard deviation of Recursive Shadowcasting, as if the FOV source is not in a pathway, but in one of the four clusters of rectangles, then there will be relatively few visible cells. Both Rectangle FOV and FOV update perform similarly to Environment 3 here. This makes sense as Environment 4 also involves many clustered rectangles which may occlude each other.

## 5. CONCLUSION

In this paper we presented two new algorithms for calculating Field of Vision over 2D grids, with the goal of making FOV calculation feasible at high grid sizes. Rectangle FOV accomplishes this by representing vision-blocking terrain in a compact and efficient manner. FOV Update uses this compact representation and a previously calculated FOV to calculate a new FOV with a minimal number of cell visibility status assignments. We then compared these algorithms to Recursive Shadowcasting, the previously fastest FOV algorithm. From this experimental evaluation, we made the following observations:

Our algorithms address the deficiencies of Resursive Shadowcasting when many cells are visible; however, our algorithms have limitations of their own when few cells are visible or when an FOV has not yet been calculated. When calculating an FOV from scratch, Recursive Shadowcasting performs best at low grid sizes and enclosed environments, while Rectangle FOV performs best at high grid sizes. Our FOV update algorithm is superior to the other two algorithms at medium and high grid sizes, but an FOV must be calculated first. Because of this, there is no universal best FOV algorithm.

Based on our experimental results, we recommend the use of a combination of algorithms if fastest computation of the FOV is desired:

- An FOV should be updated using **FOV Update** when grid size is above 512*512. At lower grid sizes, or when FOV needs to be calculated from scratch, one of the below algorithms should be used.

- An FOV should be calculated with **Recursive Shadowcasting** when grid size is 512*512 or lower, or when the environment is very "enclosed" (e.g. Environment 1).

- An FOV should be calculated with **Rectangle FOV** when grid size is above 512*512 and the current environment is not very "enclosed" (e.g. Environments 2-4).

However, not all implementors will care about minimizing average running time and would perhaps prefer to minimize the chance that FOV calculation takes long enough to be problematic. Recall that, as discussed at the end of Section 1.1, computer games generally compute a new display frame roughly every 17ms, though this time can be as low as 4ms on a modern display. FOV does not need to be calculated for every frame, but its computation must be fast enough to not delay frame rendering or starve other game processes of system resources.

Our algorithms have much more consistent performance than Recursive Shadowcasting due to their running time not being as strongly affected by the number of visible cells. Recursive Shadowcasting's highest mean running time was 46ms (with a very high standard deviation), Rectangle FOV's was 10ms, and FOV Update's was 1.8ms. Therefore, if an implementor would prefer to ensure that the running time of the FOV algorithm is always adequate, our algorithms can be used in all cases. The initial FOV calculation can be performed with Rectangle FOV, and subsequent FOV calculation can be performed with FOV Update in. Additionally, the initial FOV calculation can occur while the game is loading an environment and displaying a loading screen, which would eliminate the possibility for Rectangle FOV's running time to cause issues.

When evaluating these algorithms we tried to not make assumptions about how a particular game may use the FOV, so there is room for further improvements when certain assumptions are made. For example, the running times of Rectangle FOV and FOV Update might be improved in games with environments similar to Environment 1 or Environment 4 by using portal-based occlusion culling. Portal-based culling would allow these algorithms to determine which portions of a more "structured" environment are not visible, which would allow the algorithms to skip many visibility checks and avoid resetting the entire FOV grid. Portal based culling is likely to not be effective in less "structured" environments such as Environment 2 or Environment 3, where the added overhead might result in increased running time.

**ACKNOWLEDGEMENTS**

Authors partially supported by the Natural the Natural Sciences and engineering Research Council of Canada grants 04667-2015 RGPIN and 06423-2020 RGPIN